\documentclass{article}
\usepackage[main,final]{neurips_2026}

\usepackage[utf8]{inputenc}
\usepackage[T1]{fontenc}
\usepackage{hyperref}
\usepackage{url}
\usepackage{booktabs}
\usepackage{amsmath}
\usepackage{amsfonts}
\usepackage{amssymb}
\usepackage{nicefrac}
\usepackage{microtype}
\usepackage{xcolor}
\usepackage{multirow}
\usepackage{colortbl}
\usepackage{arydshln}
\usepackage{graphicx}
\usepackage{algorithm}
\usepackage{hyperref}
\usepackage{algpseudocode}
\usepackage[inline]{enumitem}
\usepackage{subcaption}
\usepackage{makecell}
\definecolor{mygray}{gray}{0.9}
\definecolor{mygray}{gray}{0.9}
\definecolor{myhl}{gray}{0.92}
\definecolor{mycatgray}{gray}{0.90}
\newcommand{\up}[1]{{\color{green!55!black}(+#1)}}

\newcommand{\spd}[1]{\textcolor{green!55!black}{#1$\times$}}

\title{Rethinking in Spikes: Mitigating Hallucinations in MDLMs with Step-Aware Decoding}

\author{
\begin{tabular}{@{}c@{}}
\textbf{Zhongxing Xu}$^{1}$\thanks{
Equal contribution.\url{https://mdlm-steprefine.github.io/}.
} \quad
\textbf{Zhonghua Wang}$^{1*}$ \quad
\textbf{Zhe Qian}$^{1}$ \quad
\textbf{Shiyan Su}$^{1}$ \quad
\textbf{Ming Hu}$^{1}$ \\[2pt]
\textbf{Xiaocheng Zou}$^{4}$ \quad
\textbf{Wei Feng}$^{1}$ \quad
\textbf{Mingquan Lin}$^{5}$ \quad
\textbf{Yuyin Zhou}$^{3}$ \\[2pt]
\textbf{Yifan Peng}$^{2}$ \quad
\textbf{Hamid Rezatofighi}$^{1}$ \quad
\textbf{Zongyuan Ge}$^{1}$
\end{tabular}
\\[4pt]
$^1$ Monash University \quad
$^2$ Cornell University \quad
$^3$ UC Santa Cruz
\\[2pt]
$^4$ Northeastern University \quad
$^5$ University of Minnesota
\\[3pt]
\small
\texttt{\{zhongxing.xu,zhonghua.wang\}@monash.edu}
}

\begin{document}

\maketitle

\begin{abstract}
Recent advancements in multimodal diffusion language models (MDLMs) have exhibited strong global modeling capabilities through iterative refinement. We observe that mask uncertainty follows an overall downward trend during iterative denoising, while a few denoising steps exhibit localized entropy spikes closely associated with hallucinations. We argue that  anomalous steps can be identified by measuring the entropy deviation of each denoising step from its neighboring temporal window. We hypothesize that discrete token commitment discards the distributional information at masked positions, limiting the continued interaction between candidate semantics and the context during entropy spikes. With this goal, we present \textbf{StepRefine}, a plug-and-play decoding strategy that leverages semantic context to achieve reliable denoising. The core of our method lies in a step-aware switch between discrete denoising and latent refinement. The model employs probability-weighted embeddings at entropy spike steps to preserve diverse reasoning hypotheses and performs multiple rounds of continuous refinement. Meanwhile, a refinement controller monitors distributional stability to reduce unnecessary loops and switch the model back to standard discrete denoising. Through extensive experiments, StepRefine demonstrates significant hallucination mitigation across different MDLMs on multimodal benchmarks.
\end{abstract}

\section{Introduction}

Diffusion language models~\citep{li2022diffusion,zhang2023planner,han2023ssd} replace autoregressive left-to-right generation with iterative denoising, forming an alternative paradigm for sequence generation. Current diffusion language models typically adopt a discrete masked diffusion mechanism~\citep{nie2025large,ye2025dream,zhu2025lladamoe,shi2024simplified}, in which multiple masked positions are predicted in parallel and progressively committed to discrete tokens. This paradigm has recently been extended to multimodal settings. Multimodal diffusion language models (MDLMs)~\citep{you2025lladav,you2026llada,li2025lavida,ye2025dreamvl,tian2025mmada} leverage bidirectional attention over visual and textual inputs to model global context, demonstrating strong potential in perception and reasoning tasks.

Despite these advantages, MDLMs remain susceptible to hallucination. Existing inference-time methods for diffusion language models primarily focus on efficiency, such as cache reuse~\citep{jiang2025d,wudynamic,zhu2026dllm} and denoising scheduling~\citep{lu2025adablock,liu2026ni,kim2025rainbow}, but these methods typically do not improve sequence reliability. Other decoding strategies attempt to improve quality by modifying token commitment policies, for example, by re-masking unreliable tokens~\citep{hong2025wide,huang2025don} or introducing continuous token representations~\citep{zhong2026beyond,hersche2025soft,chen2026dmax}. 
However, these methods typically treat uncertainty uniformly across denoising steps and do not explicitly account for temporal irregularities in the denoising trajectory.

In this paper, we examine hallucination in MDLMs from the perspective of step-wise uncertainty dynamics across denoising steps. A natural expectation is that predictive uncertainty should steadily decrease as the sequence becomes progressively less masked and the context becomes more complete. Our analysis shows that the average entropy of the masked sequence indeed decreases. However, this decrease is not monotonic. A small number of steps show sharp entropy increases relative to their local temporal window. We refer to these steps as entropy spikes.

Through statistical analysis, entropy spike steps are rare but highly correlated with hallucinated tokens in the final response. We observe that entropy spike steps exhibit a non-uniform distribution along the denoising trajectory, mainly occurring in the early-to-middle stages. At these stages, a large portion of the sequence remains masked, leaving the context information still limited. Moreover, we find that replacing standard confidence-based decoding with random decoding at entropy spike steps causes little performance change, whereas applying the same perturbation at ordinary steps results in clear degradation. At entropy spike steps, confidence signals may become less discriminative, making discrete decoding less reliably grounded in the current context. Therefore, our findings suggest that identifying entropy spike steps and applying contextual refinement before discrete commitment is crucial to mitigating hallucinations.

In this work, we propose StepRefine, a plug-and-play decoding strategy that leverages dense contextual cues to improve denoising reliability. StepRefine adaptively switches between standard discrete denoising and latent probabilistic refinement based on the entropy dynamics of each step from its local temporal window. Specifically, StepRefine computes the average mask entropy at each denoising step and detects entropy spikes by comparing the current entropy change against trends within a local window. When such spikes are detected, the model switches to a probabilistic refinement mode, constructs probability-weighted token embeddings that implicitly preserve diverse hypotheses, and promotes sufficient interaction between candidate semantics and the context through iterative refinement. In addition, a refinement controller monitors distributional stability to reduce unnecessary iterations and switch the model back to standard discrete denoising when appropriate. 

Extensive experiments show that StepRefine significantly reduces hallucinations across multiple MDLM models on diverse reasoning benchmarks, demonstrating its effectiveness. Our main contributions are as follows:

\begin{itemize}
    \item We analyze the relationship between entropy spike steps and hallucinations in multimodal diffusion models from the perspective of uncertainty dynamics across denoising steps.
    
    \item We propose StepRefine, a plug-and-play decoding approach that mitigates hallucinations through entropy spike detection and step-aware mode switching.
    
    \item Extensive evaluations across four MDLM backbones and diverse multimodal benchmarks show the superior performance of StepRefine, offering an effective solution for multimodal diffusion hallucinations. 
\end{itemize}

\section{Preliminary}

An MDLM takes an image and a textual prompt as input and generates text through iterative denoising. Starting from a sequence of all mask tokens, the model predicts candidate tokens for all masked positions in parallel at each step. A remasking strategy then commits a subset of these candidates as final tokens while resetting the rest to masked for the next step, until all positions are finalized. The raw image is encoded and projected into a sequence of vision tokens $\mathbf{v}$, while the textual prompt is tokenized into a sequence of text tokens $\mathbf{q}$. The generation sequence is initialized as $\mathbf{x}_T = [\mathrm{MASK}]^{L}$, where $L$ is a predefined generation budget, $[\mathrm{MASK}]$ is a special mask token from the vocabulary $\mathcal{V}$, and $T$ is the total number of denoising steps. At denoising step $t \in \{T, T-1, \ldots, 1\}$, the complete multimodal input sequence is $\mathbf{s}_t = \mathbf{v} \oplus \mathbf{q} \oplus \mathbf{x}_t$, where $\mathbf{x}_t = (x_t^1, x_t^2, \ldots, x_t^L)$ is the generation sequence at step $t$ and $x_t^i$ is the token at position $i$.

\textbf{Discrete Decoding.}
Let $\mathcal{M}_t=\{i \mid x_t^i=[\mathrm{MASK}]\}$ denote the set of masked positions at step $t$. At each denoising step, the MDLM $D_\theta$ takes $\mathbf{s}_t$ as input and predicts per-position token distributions over all positions in parallel. Discrete greedy decoding then selects the most likely tokens for each masked position:
\begin{equation}
\mathbf{p}_t = D_\theta(\mathbf{s}_t) \in \Delta(\mathcal{V})^{L},
\qquad
\hat{x}_t^i = \arg\max_{v\in\mathcal{V}} \mathbf{p}_t^i(v), \quad i\in\mathcal{M}_t,
\label{eq:discrete}
\end{equation}
where $\Delta(\mathcal{V})^L$ denotes a sequence of $L$ probability simplices over $\mathcal{V}$, and $\mathbf{p}_t^i$ is the prediction distribution at position $i$. Discrete decoding produces a candidate token $\hat{x}_t^i$ at each masked position, while the remasking step determines which candidates are actually committed.

\begin{figure}[t]
  \centering  
  \includegraphics[width=1\linewidth]{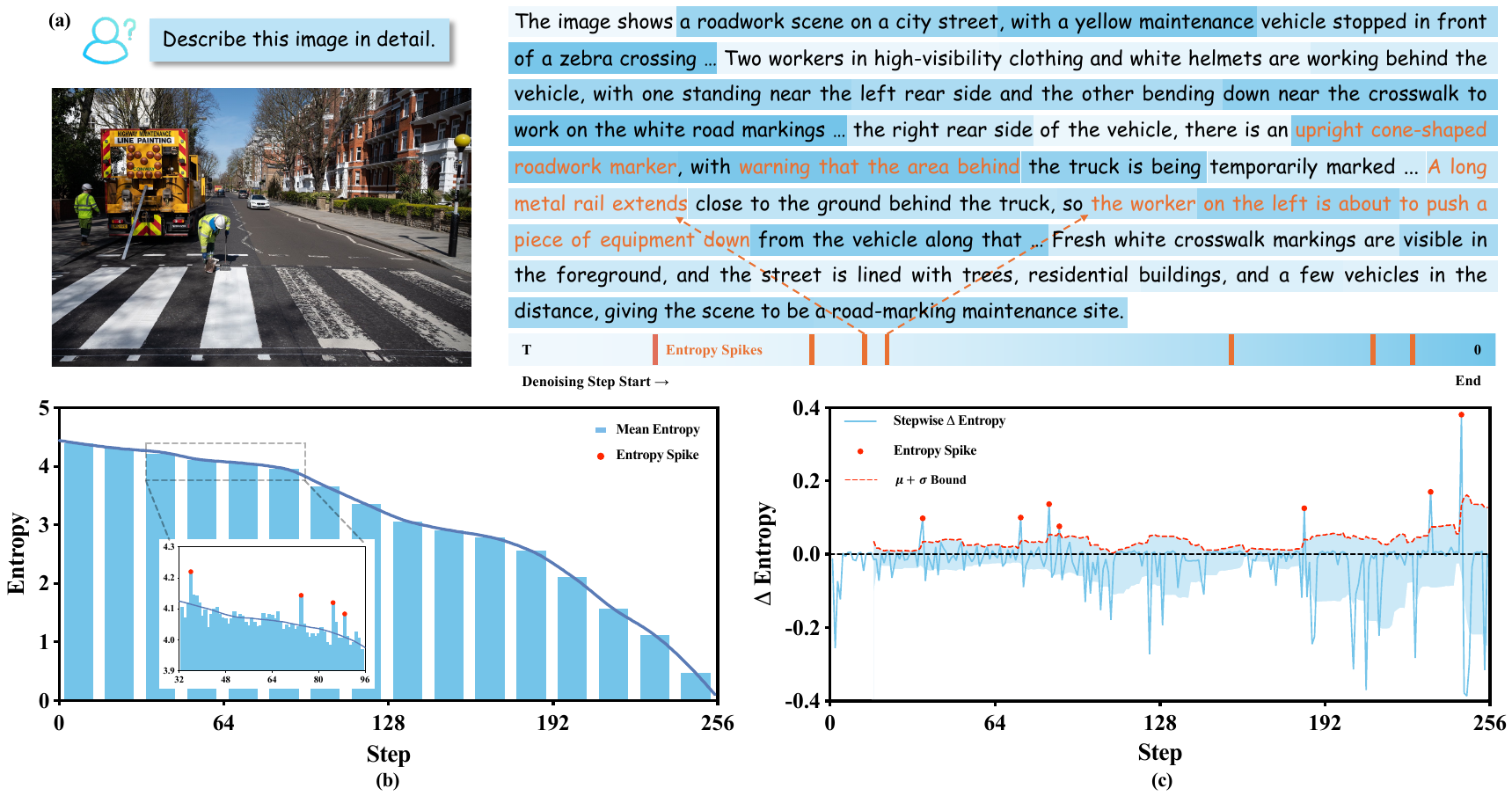}
  \caption{Entropy spikes in denoising dynamics. (a) A representative example illustrating the correlation between entropy spikes and hallucinations. Tokens committed at steps with an entropy spike closely correspond to hallucinated content in the response. (b) Mean mask entropy decreases steadily across denoising, while a few steps show sharp rises that break the local trend, which we refer to as entropy spikes. (c) A step is flagged as an entropy spike when its stepwise entropy change exceeds the $\mu+\sigma$ bound from the calibration window. }
  \label{fig:main1}
\end{figure}

\vspace{-0.1cm}
\textbf{Latent Refinement.}
Let $\mathbf{e}(v) \in \mathbb{R}^d$ denote the embedding of token $v$, where $d$ is the embedding dimension. Discrete decoding collapses the predictive distribution $\mathbf{p}_t^i$ into a single token, discarding its distributional information. Latent refinement instead converts the prediction distribution into a probability-weighted embedding and further refines it in the continuous representation space:
\begin{equation}
\tilde{\mathbf{e}}_{t}^{i} = \mathbb{E}_{v \sim \mathbf{p}_t^i}\left[\mathbf{e}(v)\right] \in \mathbb{R}^d, \quad i \in \mathcal{M}_t,
\label{eq:latent}
\end{equation}
where $\mathbb{E}$ denotes the expectation under the distribution $\mathbf{p}_t^i$. The resulting soft embedding $\tilde{\mathbf{e}}_{t}^{i}$ replaces the $[\mathrm{MASK}]$ embedding at position $i$ in the next forward pass, allowing candidate semantics to further interact with the context before discrete token decoding.

\textbf{Remasking.}
Although discrete decoding produces candidate tokens for all masked positions at each step, only a subset of $\hat{x}_t^i$ is committed into the generation sequence. A remasking strategy selects an update set $\mathcal{U}_t\subseteq\mathcal{M}_t$, where the chosen positions are committed and the rest stay masked. Common choices include random sampling that selects positions uniformly, low-confidence remasking that commits the highest-confidence positions according to a step-wise schedule on $|\mathcal{U}_t|$, and top-$k$ marginal selection that commits a fixed number of $k$ positions with the largest marginal probability per step. The generation sequence is then updated by committing selected positions while leaving the others unchanged:
\begin{equation}
x_{t-1}^i =
\begin{cases}
\hat{x}_t^i, & i\in\mathcal{U}_t,\\
x_t^i, & i\notin\mathcal{U}_t.
\end{cases}
\label{eq:remask}
\end{equation}
The masked-position set is accordingly updated as $\mathcal{M}_{t-1}=\mathcal{M}_t\setminus\mathcal{U}_t$. Through repeated prediction and remasking, the model progressively refines an initially fully masked sequence $\mathbf{x}_T$ into a fully decoded sequence $\mathbf{x}_0$.

\begin{figure}[t]
  \centering  
  \includegraphics[width=1\linewidth]{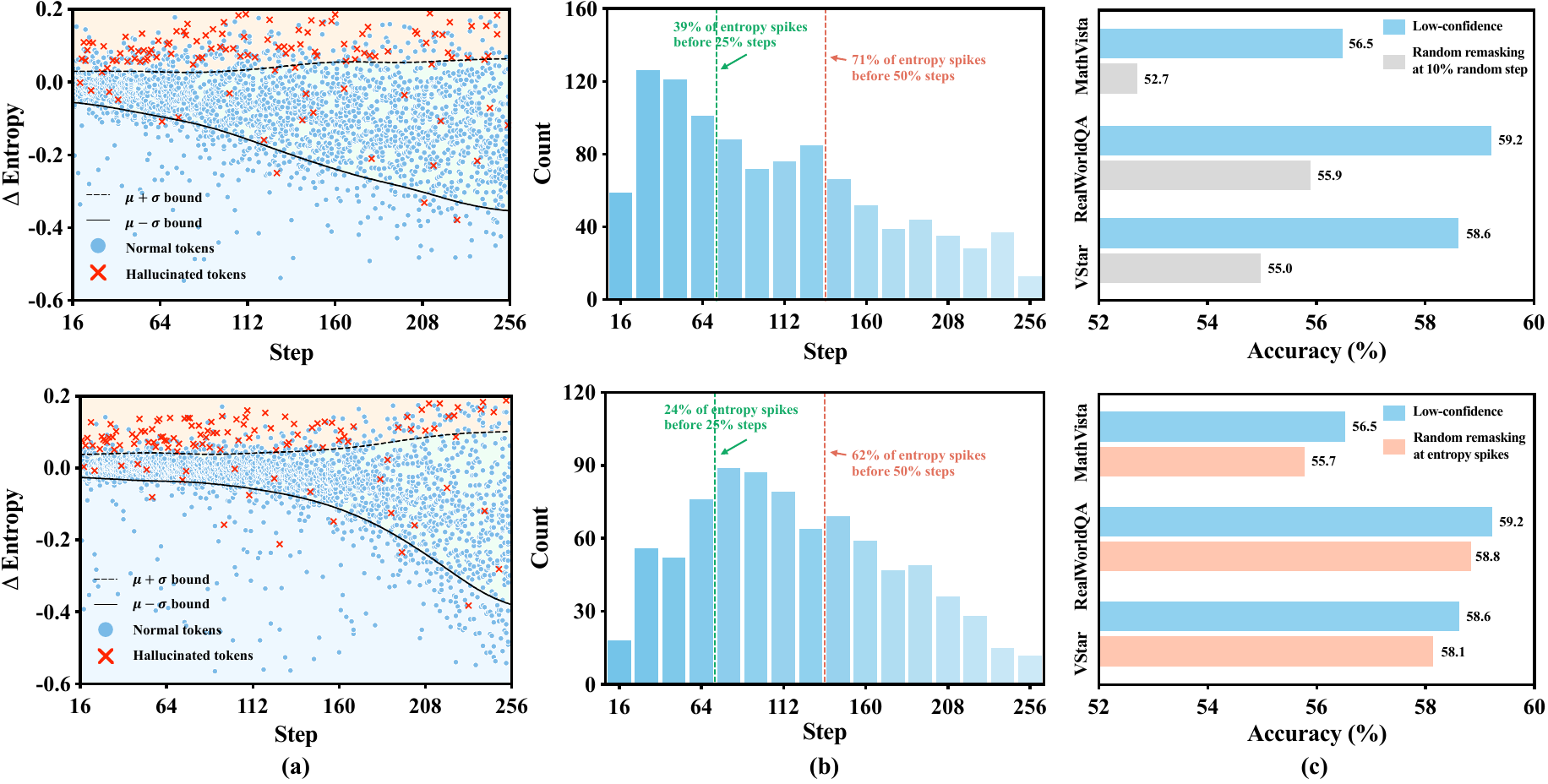}
\caption{Entropy spikes characterization across decoding steps. (a) Token-level scatter of stepwise entropy change on LLaDA-V (top) and Dream-VL (bottom). Roughly 80\% of hallucinated tokens fall in the area above $\mu+\sigma$. (b) Distribution of entropy spikes across denoising steps on LLaDA-V (top) and Dream-VL (bottom). On both models, entropy spikes mostly occur in the early-to-middle decoding stage, when few tokens have been committed and the available context is still limited. (c) Effect of random remasking on performance using LLaDA-V. Top: replacing low-confidence remasking with random remasking at 10\% of randomly selected steps causes a substantial drop on three datasets. Bottom: applying the same random remasking only at entropy spike steps yields negligible change. At entropy spike steps, confidence signals may become less discriminative, making discrete commitment decisions less reliably grounded in the current contextual information.}

\vspace{-0.1cm}
  \label{fig:main2}
\end{figure}

\section{Methodology}

We propose StepRefine, a step-aware decoding strategy for mitigating hallucinations in
MDLMs. Section~\ref{sec:3.1} presents entropy spike detection, along with an analysis of entropy spikes. Section~\ref{sec:3.2} introduces step-aware refinement switching with adaptive termination. For clarity, Algorithm~\ref{alg:steprefine} provides the pseudocode for the decoding process of StepRefine.

\subsection{Entropy Spikes Detection}
\label{sec:3.1}
During iterative denoising, the mean mask entropy over masked positions decreases as the model gradually commits tokens (Fig.~\ref{fig:main1}b). However, the decrease is not monotonic. A small number of steps exhibit sharp local increases that break the trend, which we refer to as entropy spikes. 
We measure this entropy at each denoising step $t$, defined as the expected Shannon entropy of $\mathbf{p}_t^i$ over masked positions:
\begin{equation}
\bar{H}_t = \mathbb{E}_{i \sim \mathbf{U}(\mathcal{M}_t)}\!\left[\,-\sum_{v\in\mathcal{V}} \mathbf{p}_t^i(v)\log \mathbf{p}_t^i(v)\,\right],
\label{eq:entropy}
\end{equation}
where $\mathbf{U}(\mathcal{M}_t)$ denotes the uniform distribution over the masked positions $\mathcal{M}_t$. A step is then classified as an entropy spike if its entropy change exceeds the local upper bound:
\begin{equation}
\Delta\bar{H}_t \triangleq \bar{H}_t - \bar{H}_{t+1}, \qquad
\mathcal{T}_{\mathrm{spike}} \triangleq \bigl\{\, t \,\mid\, \Delta\bar{H}_t > \mu_t + \sigma_t \,\bigr\},
\label{eq:spike}
\end{equation}
where $\mu_t$ and $\sigma_t$ are the mean and standard deviation of the $w$ preceding stepwise entropy changes along the denoising trajectory, computed over $\{\Delta\bar{H}_{t+1}, \ldots, \Delta\bar{H}_{t+w}\}$. As shown in Fig.~\ref{fig:main1}c, this criterion captures steps where mean mask entropy deviates sharply from the local denoising trend.

\vspace{-0.2cm}

\paragraph{Entropy Spike Analysis.} Figure~\ref{fig:main2}(a) demonstrates that entropy spike steps are relatively rare along the denoising trajectory, while approximately 80\% of hallucinated tokens are located within the entropy spike region. Fig.~\ref{fig:main2}(b) further shows the distribution of entropy spikes across the denoising process. Entropy spikes are mainly concentrated in the early-to-middle stages: about 71\% occur within the first 50\% of steps in LLaDA-V, while about 62\% occur within the first 50\% of steps in Dream-VL. This suggests that entropy spikes are non-uniformly distributed and are more likely to occur when many positions remain masked and the available context is still insufficient. Low-confidence remasking is an effective deterministic strategy~\citep{nie2025large}, which preferentially preserves high-confidence tokens. Fig.~\ref{fig:main2}(c) shows that applying the same random-remasking perturbation causes a clear performance drop at randomly selected 10\% steps, but only a minor change at entropy spike steps. At entropy spike steps, confidence signals may become less discriminative, making discrete commitment decisions less reliably grounded in the current contextual information.

\begin{algorithm}[t]
\small
\caption{StepRefine}
\label{alg:steprefine}
\begin{algorithmic}[1]
\Require MDLM $D_\theta$, embedding matrix $\mathbf{W}_{\text{emb}}$, denoising budget $T$, generation length $L$, window size $w$, stability threshold $\epsilon$, prompt $\mathbf{x}_{\text{prompt}}$
\State Initialize $\mathbf{x}_T \gets \text{concat}(\mathbf{x}_{\text{prompt}},[\mathrm{MASK}]^{L})$,\ \ $\mathbf{E}_T \gets \text{Embed}(\mathbf{x}_T)$,\ \ $t \gets T$
\While{$t \geq 1$}
    \State $\mathbf{p}_t \gets D_\theta(\mathbf{E}_t)$
    \State Compute $\bar{H}_t,\ \Delta\bar{H}_t,\ \mu_t,\ \sigma_t$,\ \ $k \gets 0$ \Comment{forward pass + spike detection}
    \If{$\Delta\bar{H}_t > \mu_t + \sigma_t$} \Comment{$t \in \mathcal{T}_{\mathrm{spike}}$}
        \State $\tilde{\mathbf{p}}_{t,0} \gets \mathbf{p}_t$
        \Repeat
            \State $k \gets k + 1$,\ \ $\mathbf{E}_t[\mathcal{M}_t] \gets \tilde{\mathbf{p}}_{t,k-1}[\mathcal{M}_t]\,\mathbf{W}_{\text{emb}}$,\ \ $\tilde{\mathbf{p}}_{t,k} \gets D_\theta(\mathbf{E}_t)$ \Comment{StepRefine iteration}
        \Until{$\overline{D}_{\mathrm{KL}}^{(k)} < \epsilon$} \Comment{exit on distributional stability}
        \State $\mathbf{p}_t \gets \tilde{\mathbf{p}}_{t,k}$
    \EndIf
    \State Decode $\hat{x}_t^i \gets \arg\max_{v\in\mathcal{V}} \mathbf{p}_t^i(v)$ for $i\in\mathcal{M}_t$, select $\mathcal{U}_t \subseteq \mathcal{M}_t$
    \State Update $\mathbf{x}_{t-1}$,\ \ $\mathbf{x}_{t-1-k} \gets \mathbf{x}_{t-1}$,\ \ $t \gets t-1-k$,\ \ $\mathbf{E}_t \gets \text{Embed}(\mathbf{x}_t)$ \Comment{commit, then skip $k$ indices}
\EndWhile
\State \Return $\mathbf{x}_0$
\end{algorithmic}
\end{algorithm}

\subsection{Step-Aware Mode Switch}
\label{sec:3.2}

StepRefine adopts a step-aware switching strategy that alternates between standard discrete decoding and latent refinement, gated by the spike detection criterion (Eq.~\ref{eq:spike}). At each denoising step $t$, it first performs a standard forward pass with $\mathbf{e}([\mathrm{MASK}])$ as input at masked positions to obtain the predictive distribution $\mathbf{p}_t^i$ for $i \in \mathcal{M}_t$ and the corresponding spike indicator. 
At ordinary steps $t \notin \mathcal{T}_{\text{spike}}$, it follows the standard discrete decoding path. 
At entropy spike steps $t \in \mathcal{T}_{\mathrm{spike}}$, StepRefine activates latent refinement, which performs iterative forward passes within the same denoising step before discrete decoding. 
Let $\tilde{\mathbf{p}}_{t,0}^i \triangleq \mathbf{p}_t^i$ denote the initial distribution from the standard forward pass.
At each refinement iteration $k \geq 1$, StepRefine constructs a probability-weighted embedding $\tilde{\mathbf{e}}_{t,k}^i$ from the previous distribution $\tilde{\mathbf{p}}_{t,k-1}^i$ and replaces the $[\mathrm{MASK}]$ embedding as input to the next forward pass, which produces an updated distribution $\tilde{\mathbf{p}}_{t,k}^i$. Formally,
\begin{equation}
\tilde{\mathbf{e}}_{t,k}^i = \mathbb{E}_{v \sim \tilde{\mathbf{p}}_{t,k-1}^i}\!\left[\mathbf{e}(v)\right], \quad i \in \mathcal{M}_t,\ k = 1, 2, \ldots.
\label{eq:steprefine_iter}
\end{equation}
Together with the forward pass, Eq.~\ref{eq:steprefine_iter} forms a recursive refinement loop between $\tilde{\mathbf{p}}_{t,k}^i$ and $\tilde{\mathbf{e}}_{t,k}^i$ at each masked position. It carries distributional information about candidate tokens into subsequent forward passes, letting the soft embedding interact with the visual-textual context before discrete decoding.

To prevent unnecessary refinement, the refinement controller monitors the change between consecutive predictive distributions. At iteration $k \geq 1$, we compute the change as the average Kullback-Leibler (KL) divergence ($D_{\mathrm{KL}}(\cdot\,\|\,\cdot)$) at masked positions:
\begin{equation}
\overline{D}_{\mathrm{KL}}^{(k)} = \mathbb{E}_{i \sim \mathbf{U}(\mathcal{M}_t)}\!\left[\,D_{\mathrm{KL}}\!\left(\tilde{\mathbf{p}}_{t,k}^{i}\,\|\,\tilde{\mathbf{p}}_{t,k-1}^{i}\right)\,\right].
\end{equation}
The iteration terminates once $\overline{D}_{\mathrm{KL}}^{(k)} < \epsilon$, where $\epsilon > 0$ is a stability threshold. The final refined distribution $\tilde{\mathbf{p}}_{t,k}^i$ is then used for standard discrete decoding and remasking to commit tokens and obtain $\mathbf{x}_{t-1}$. To maintain a fixed inference budget, the additional forward passes for refinement are counted against the total denoising budget of $T$. If $k$ refinement passes are used at step $t$, the next iteration begins at index $t-1-k$ rather than $t-1$. Thus, the total number of forward passes remains the same.

\section{Experiments}

\subsection{Experimental Settings}
\textbf{Baselines and Evaluation Benchmarks.} We evaluate StepRefine on four representative MDLMs (LLaDA-V-8B~\citep{you2025lladav}, LLaDA-O-8B~\citep{you2026llada}, Dream-VL-7B~\citep{ye2025dreamvl}, and LaViDa-L-8B~\citep{li2025lavida}). 
We evaluate on a diverse suite of multimodal benchmarks covering four general and domain-specific reasoning categories: (i) General reasoning \& understanding on VStar~\citep{wu2024v}, MMVP~\citep{tong2024eyes}, MMEval-Pro~\citep{huang2025mmevalpro}, VMCBench~\citep{zhang2025automated}, RealWorldQA~\citep{grok2beta2024}; (ii) General hallucination assessment on MMHalu~\citep{sun2023aligning}, Bingo~\citep{cui2023holistic}, POPE~\citep{li2023pope}, (iii) Mathematical reasoning on MathVision~\citep{wang2024mathvision}, MathVista~\citep{lu2023mathvista}; and (iv) Scientific reasoning on the MMK12~\citep{meng2025mm} Physics and Chemistry subsets.

\textbf{Implementation Details.} Across all backbones and benchmarks, we decode with generation length $L=256$, total denoising steps $T=256$, and low-confidence remasking. The stability threshold $\epsilon$ is set to 0.01. Details of the compared methods are provided in Appendix~\ref{appendix:baseline}. 

\subsection{Main Results}

\begin{table*}[!t]
    \centering
\caption{Benchmark results of StepRefine with different MDLMs across four benchmark categories. Scores are reported for MMHalu (0--6) and Bingo (1--5); accuracy is reported for all other benchmarks.}
\label{tab:main_results_general_v1}
    \small
    \setlength{\tabcolsep}{1.4pt}
    \renewcommand{\arraystretch}{1.6}
    \begin{tabular}{lcccc>{\columncolor{myhl}}cc>{\columncolor{myhl}}cc>{\columncolor{myhl}}cc>{\columncolor{myhl}}c}
        \toprule
        \multirow{2}{*}{\textbf{Benchmark}}
        & \multicolumn{5}{c}{\textbf{LLaDA-V-8B}}
        & \multicolumn{2}{c}{\textbf{LLaDA-O-8B}}
        & \multicolumn{2}{c}{\textbf{Dream-VL-7B}}
        & \multicolumn{2}{c}{\textbf{LaViDA-L-8B}} \\
        \cmidrule(lr){2-6} \cmidrule(lr){7-8} \cmidrule(lr){9-10} \cmidrule(lr){11-12}
        & Base & CoTA & ReMix & WINO & \cellcolor{myhl}\textbf{StepRefine}
        & Base & \textbf{StepRefine}
        & Base & \textbf{StepRefine}
        & Base & \textbf{StepRefine} \\
        \midrule
        \addlinespace[-0.3pt]
        \multicolumn{12}{c}{\textit{\textbf{General Reasoning \& Understanding}}} \\
        \addlinespace[-1pt]
        \midrule
        VStar         & 55.5 & 57.1 & 56.5 & 56.5 & \textbf{58.6}~\up{3.1} & 57.1 & \textbf{58.1}~\up{1.0} & 69.1 & \textbf{72.8}~\up{3.7} & 48.7 & \textbf{51.3}~\up{2.6} \\
        MMVP          & 40.7 & 42.7 & 43.3 & 44.0 & \textbf{45.3}~\up{4.6} & 44.7 & \textbf{47.3}~\up{2.6} & 32.0 & \textbf{35.3}~\up{3.3} & 42.3 & \textbf{46.7}~\up{4.4} \\
        MMEval-Pro    & 68.4 & 67.9 & 63.5 & 68.5 & \textbf{71.6}~\up{3.2} & 57.9 & \textbf{60.8}~\up{2.9} & 68.5 & \textbf{70.9}~\up{2.4} & 55.3 & \textbf{59.8}~\up{4.5} \\
        VMCBench      & 73.1 & 73.8 & 71.0 & 74.7 & \textbf{74.4}~\up{1.3} & 69.8 & \textbf{73.0}~\up{3.2} & 71.8 & \textbf{73.6}~\up{1.8} & 61.5 & \textbf{63.5}~\up{2.0} \\
        RealWorldQA   & 57.5 & 58.3 & 58.8 & 58.6 & \textbf{59.2}~\up{1.7} & 56.0 & \textbf{57.3}~\up{1.3} & 64.1 & \textbf{65.5}~\up{1.4} & 49.2 & \textbf{52.8}~\up{3.6} \\
        \midrule
        \addlinespace[-0.3pt]
        \multicolumn{12}{c}{\textit{\textbf{Hallucination Benchmarks}}} \\
        \addlinespace[-1pt]
        \midrule
        MMHalu        & 3.35 & 3.39 & 3.25 & 3.43 & \textbf{3.50}~\up{2.5} & 3.76 & \textbf{4.01}~\up{4.2} & 3.54 & \textbf{3.68}~\up{2.3} & 2.38 & \textbf{2.54}~\up{2.7} \\
        Bingo         & 2.71 & 2.93 & 2.65 & 2.96 & \textbf{2.93}~\up{4.4} & 2.90 & \textbf{3.08}~\up{3.6} & 3.31 & \textbf{3.46}~\up{3.0} & 2.35 & \textbf{2.56}~\up{4.2} \\
        POPE-R        & 89.6 & 89.4 & 90.3 & 89.8 & \textbf{91.5}~\up{1.9} & 85.3 & \textbf{86.1}~\up{0.8} & 82.6 & \textbf{85.7}~\up{3.1} & 66.8 & \textbf{69.1}~\up{2.3} \\
        \midrule
        \addlinespace[-0.3pt]
        \multicolumn{12}{c}{\textit{\textbf{Mathematical Reasoning}}} \\
        \addlinespace[-1pt]
        \midrule
        MathVision    & 25.6 & 27.3 & 26.4 & 28.8  & \textbf{29.1}~\up{3.5} & 20.8 & \textbf{23.7}~\up{2.9} & 26.7 & \textbf{30.9}~\up{4.2} & 26.1 & \textbf{28.9}~\up{2.8} \\
        MathVista     & 54.7 & 55.2 & 54.6 & 55.9  & \textbf{56.5}~\up{1.8} & 64.8 & \textbf{67.1}~\up{2.3} & 57.8 & \textbf{59.6}~\up{1.8} & 47.0 & \textbf{47.2}~\up{0.2} \\
        \midrule
        \addlinespace[-0.3pt]
        \multicolumn{12}{c}{\textit{\textbf{Scientific Reasoning}}} \\
        \addlinespace[-1pt]
        \midrule
        MMK12-Phys    & 37.6 & 38.2 & 38.8 & 38.0 & \textbf{39.4}~\up{1.8} & 24.4 & \textbf{25.2}~\up{0.8} & 35.8 & \textbf{38.0}~\up{2.2} & 33.0 & \textbf{33.4}~\up{0.4} \\
        MMK12-Chem    & 36.2 & 37.6 & 39.8 & 38.0 & \textbf{40.4}~\up{4.2} & 23.4 & \textbf{25.2}~\up{1.8} & 31.4 & \textbf{34.6}~\up{3.2} & 35.2 & \textbf{36.4}~\up{1.2} \\
        \bottomrule
    \end{tabular}
\end{table*}

\textbf{Benchmark Evaluation.} We compare StepRefine against three training-free decoding strategies for MDLMs, including CoTA~\citep{zhao2026context}, ReMix~\citep{ye2026rejection}, and WINO~\citep{hong2025wide}, as shown in Table~\ref{tab:main_results_general_v1}. On LLaDA-V-8B, StepRefine improves average accuracy on general reasoning by +2.8\% over the baseline and matches or outperforms competing decoding strategies on every general reasoning benchmark. 
It also improves on the hallucination benchmarks, yielding gains of +2.5\% on MMHalu and +4.4\% on Bingo. These improvements generalize across all backbones and benchmark categories. Every backbone shows positive gains, with average improvements ranging from +2.3\% to +2.8\%. 
These results suggest that the benefit of entropy-spike-aware refinement is robust across MDLM backbones rather than specific to any single model.
Additional results are reported in Appendix~\ref{appendix:extra_results}.

\begin{table*}[!t]
    \centering
    \caption{Compatibility of StepRefine with existing dLLM acceleration methods. We report accuracy for VStar and MMVP, score for MMHalu and Bingo, along with inference speedup and throughput (\textit{Tput}, tokens/s). Speedups are measured relative to the unaccelerated LLaDA-V baseline. }
    \label{tab:accel_compat}
    \small
    \setlength{\tabcolsep}{2pt}
    \renewcommand{\arraystretch}{1.5}
    \begin{tabular}{lcccccccccccc}
        \toprule
        \multirow{2}{*}{\textbf{Method}}
        & \multicolumn{3}{c}{\textbf{VStar}}
        & \multicolumn{3}{c}{\textbf{MMVP}}
        & \multicolumn{3}{c}{\textbf{MMHalu}}
        & \multicolumn{3}{c}{\textbf{Bingo}} \\
        \cmidrule(lr){2-4} \cmidrule(lr){5-7} \cmidrule(lr){8-10} \cmidrule(lr){11-13}
        & Acc & Speedup & Tput
        & Acc & Speedup & Tput
        & Score & Speedup & Tput
        & Score & Speedup & Tput \\
        \midrule
        LLaDA-V                   & 55.5          & \spd{1.00} & 2.75
                                  & 40.7          & \spd{1.00} & 6.61
                                  & 3.35          & \spd{1.00} & 2.75
                                  & 2.71          & \spd{1.00} & 3.33 \\
        \midrule
        Fast-dLLM                 & 52.4          & \spd{8.45} & 23.21
                                  & 37.3          & \spd{4.23} & 27.94
                                  & 3.09          & \spd{7.75} & 21.37
                                  & 2.58          & \spd{6.83} & 22.75 \\
        \textbf{\quad + StepRefine} & \textbf{56.5} & \spd{8.47} & 23.39
                                  & \textbf{42.0} & \spd{4.25} & 28.05
                                  & \textbf{3.22} & \spd{7.79} & 21.41
                                  & \textbf{2.64} & \spd{6.86} & 22.84 \\
        \midrule
        dLLM-Cache                & 53.9          & \spd{2.69} & 7.39
                                  & 42.7          & \spd{2.02} & 13.35
                                  & 3.29          & \spd{2.66} & 7.32
                                  & 2.44          & \spd{2.87} & 9.58 \\
        \textbf{\quad + StepRefine}& \textbf{57.1} & \spd{2.73} & 7.51
                                  & \textbf{44.7} & \spd{2.02} & 13.37
                                  & \textbf{3.36} & \spd{2.72} & 7.48
                                  & \textbf{2.47} & \spd{2.88} & 9.60 \\
        \midrule
        dKV-Cache                 & 54.5          & \spd{2.42} & 6.63
                                  & 42.0          & \spd{1.96} & 12.94
                                  & 3.33          & \spd{2.41} & 6.64
                                  & 2.41          & \spd{2.66} & 8.84 \\
        \textbf{\quad + StepRefine} & \textbf{56.0} & \spd{2.48} & 6.82
                                  & \textbf{43.4} & \spd{1.98} & 13.10
                                  & \textbf{3.41} & \spd{2.49} & 6.85
                                  & \textbf{2.50} & \spd{2.71} & 9.03 \\
        \bottomrule
    \end{tabular}

\end{table*}

\begin{figure}[t]
  \centering
  \vspace{1em}
  \includegraphics[width=1\linewidth]{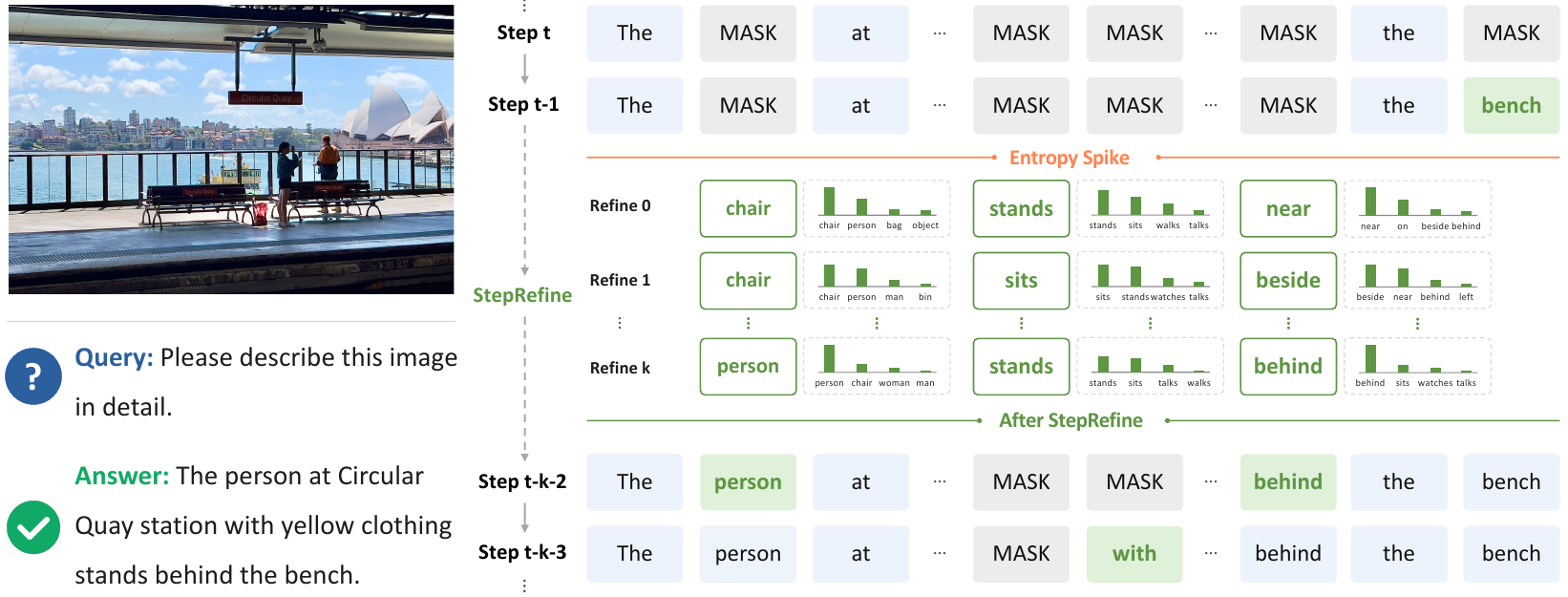}
\caption{Example visualization of StepRefine. At an entropy spike step, StepRefine reshapes the predicted distribution at masked positions, shifting the prediction toward a more context-consistent candidate, which is then committed. More qualitative examples are provided in Appendix~\ref{app:visualization}.}
\vspace{-0.1cm}
  \label{fig:visualization}
\end{figure}

\textbf{Compatibility with Acceleration Methods.} We further evaluate StepRefine in combination with three acceleration methods, including Fast-dLLM~\citep{wu2025fast}, dLLM-Cache~\citep{liu2025dllm}, dKV-Cache~\citep{ma2025dkv}. StepRefine is orthogonal to these methods by design. Each refinement round, StepRefine operates on the embedding at masked positions through standard forward passes, leaving the cache and feature-reuse schedules used by each accelerator unchanged during ordinary steps. Table~\ref{tab:accel_compat} shows that, plugged into each method, StepRefine largely maintains the speedups of these acceleration methods while improving performance by +2.9\% on VStar and +1.6\% on MMHalu. Throughput is also preserved or slightly increased after adding StepRefine, confirming that refinement does not introduce hidden latency. 
The additional speedup over each accelerator's baseline is due to the budget compensation mechanism. The $k$ refinement rounds at an entropy spike step are charged against the remaining step budget, replacing $k$ later steps rather than extending the total compute.

\textbf{Qualitative Analysis.} Figure~\ref{fig:visualization} illustrates a representative StepRefine decoding trajectory, which alternates between standard denoising steps and entropy spike steps. Within each spike step, we plot the predicted distributions at masked positions across the refinement rounds. As refinement proceeds, these distributions sharpen as the entropy of the predicted distribution decreases, and the prediction converges on the candidate that is finally committed. Notably, entropy spike steps are precisely the points where a single greedy commit is unreliable. Each refinement round feeds the soft embeddings derived from the previous distributions back into the model, letting competing candidates interact with the surrounding context.

\subsection{Ablation Study}

\begin{figure}[t]
  \centering  
  \includegraphics[width=1\linewidth]{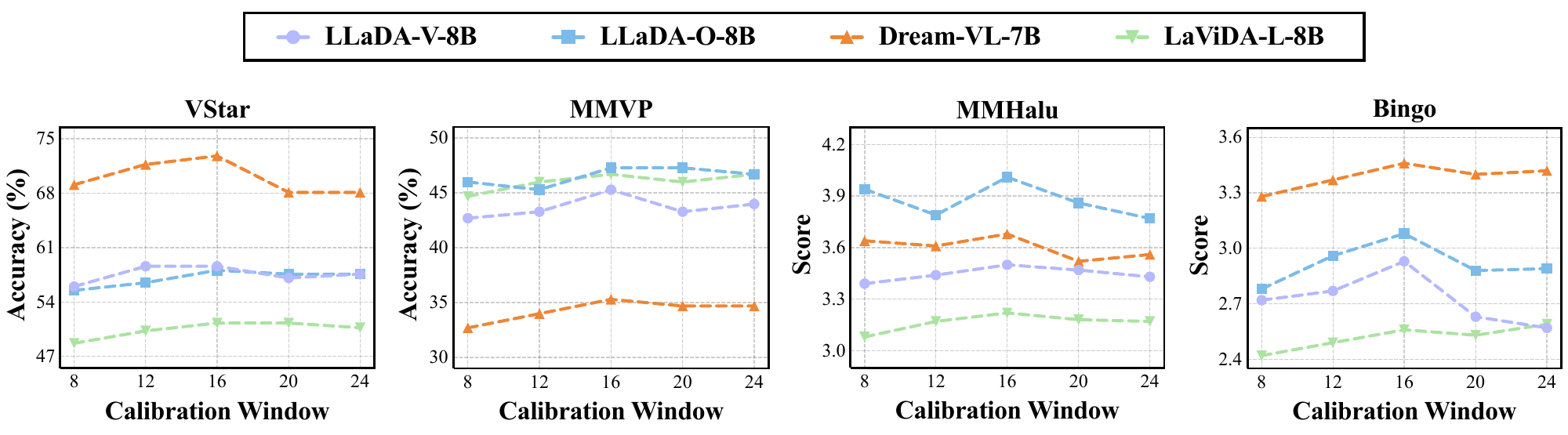}
  \vspace{-0.4cm}
  \caption{Comparisons of model performance on four datasets under different calibration window sizes. The window provides the running statistics of the entropy deviation used to detect spike steps.}
  \vspace{-1em}
  \label{fig:ablation_calibration_window}
\end{figure}

\textbf{Effect of Calibration Window Size.} We study the effect of the calibration window size $w$ used for entropy spike detection. As shown in Fig.~\ref{fig:ablation_calibration_window}, accuracy improves as the window increases up to a moderate size ($w=16$ in our experiments), and then oscillates. When the window is small, the entropy-deviation estimate is dominated by step-to-step noise and produces unstable triggers. When the window is large, entropy deviations are averaged across denoising stages with different uncertainty profiles, blurring the distinction between stable steps and steps with entropy spike. Taken together, these suggest that a moderate window provides a reasonable trade-off in our setting. 

\begin{figure}[h]
  \centering  
  \includegraphics[width=1\linewidth]{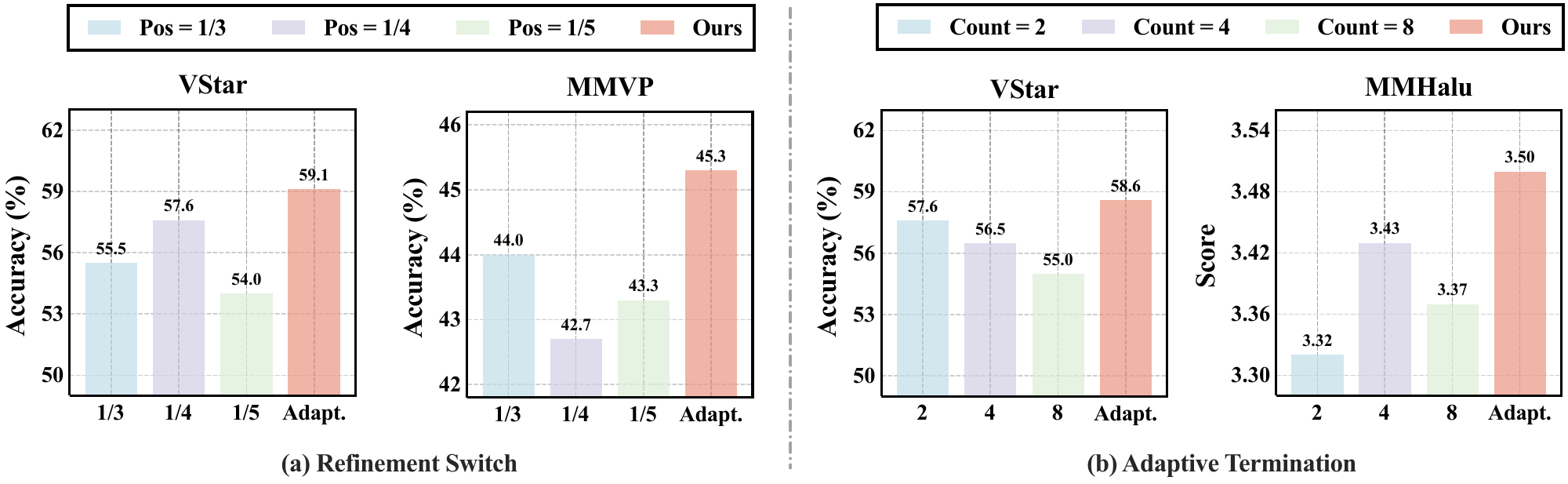}
  \vspace{-0.4cm}
\caption{Ablations on StepRefine's adaptive trigger and termination criteria. (a) Comparison between StepRefine's adaptive trigger and three fixed schedules that insert latent refinement at uniformly spaced denoising steps. (b) Comparison between StepRefine's adaptive termination criterion and fixed numbers of refinement rounds. }
  \label{fig:ablation_adaptive_trigger}
\end{figure}

\textbf{Adaptive versus Fixed Refinement Scheduling.} We compare StepRefine's adaptive trigger against three fixed schedules that insert latent refinement at uniformly spaced denoising steps. Fig.~\ref{fig:ablation_adaptive_trigger}(a) reveals that the adaptive trigger outperforms all three fixed schedules. Entropy spike steps neither follow a uniform spacing nor occur at the same positions across samples, so a periodic schedule may fail to capture spikes in some inputs while spending refinement on already-stable steps in others. In contrast, our adaptive trigger conditions both the position and frequency of refinement on local entropy deviation, applying refinement only at steps with anomalous uncertainty.

\textbf{Effect of the Refinement Controller.} We compare StepRefine's adaptive termination criterion against fixed numbers of refinement rounds. Fig~\ref{fig:ablation_adaptive_trigger}(b) shows that the adaptive scheme outperforms every fixed budget on both datasets, while no single fixed value performs consistently best across datasets. A single fixed budget allocates an equal number of refinement rounds to every spike step, regardless of whether the prediction has already converged. The adaptive criterion instead allocates refinement to each entropy spike step only while its predictive distribution continues to change, and terminates when it stabilizes.

\textbf{Robustness across Decoding Settings.}
StepRefine remains robust across diverse decoding settings. As shown in Table~\ref{tab:ablation_steprefine_combined}(a), StepRefine improves over the baseline under three remasking strategies with absolute MMHalu gains of +3.7\% to +4.5\%. Table~\ref{tab:ablation_steprefine_combined}(b) further shows consistent improvements across two generation lengths with three step budgets each, with MMHalu gains ranging from +1.8\% to +4.8\% across all six configurations. These results indicate that StepRefine generalizes across decoding configurations without per-setting tuning. The gain also holds at tight budgets, indicating that StepRefine remains effective even when total compute is constrained. 

\begin{table}[t]
    \caption{Ablations of decoding settings for LLaDA-V-8B on MMHalu. (a) Baseline vs. StepRefine across remasking strategies. (b) Baseline vs. StepRefine across step budgets and generation lengths.}
    \label{tab:ablation_steprefine_combined}
    \centering
    \begin{subtable}[t]{0.39\textwidth}
        \centering
        \small
        \caption{Remasking strategy}
        \setlength{\tabcolsep}{3pt}
        \renewcommand{\arraystretch}{1.4}
        \begin{tabular}{lcc}
            \toprule
            \textbf{Strategy} & Base & \textbf{StepRefine} \\
            \midrule
            Random          & 2.97 & \textbf{3.08} \\
            Low Confidence  & 3.35 & \textbf{3.50} \\
            Top-$k$ Margin  & 3.27 & \textbf{3.41} \\
            \bottomrule
        \end{tabular}
    \end{subtable}
    \hfill
    \begin{subtable}[t]{0.60\textwidth}
        \centering
        \small
        \caption{Step budget at two generation lengths}
        \setlength{\tabcolsep}{3pt}
        \renewcommand{\arraystretch}{1.26}
        \begin{tabular}{lcccccc}
            \toprule
            \multirow{2}{*}{\textbf{Method}}
            & \multicolumn{3}{c}{$L=256$} & \multicolumn{3}{c}{$L=128$} \\
            \cmidrule(lr){2-4} \cmidrule(lr){5-7}
            & $T{=}64$ & $T{=}128$ & $T{=}256$ & $T{=}32$ & $T{=}64$ & $T{=}128$ \\
            \midrule
            Base               & 3.13 & 3.31 & 3.35 & 3.22 & 3.30 & 3.45 \\
            \textbf{StepRefine} & \textbf{3.28} & \textbf{3.37} & \textbf{3.50} & \textbf{3.34} & \textbf{3.41} & \textbf{3.53} \\
            \bottomrule
        \end{tabular}
    \end{subtable}
\end{table}

\section{Related Work}
\textbf{Multimodal diffusion LLM.}
Diffusion large language models (dLLMs)~\citep{nie2025large,ye2025dream,zhu2025lladamoe,jo2025continuous,ai2026llada2} leverage their inherent bidirectional attention and iterative denoising process to exhibit strong global context modeling capability and flexible reasoning potential. In recent years, this paradigm has been further extended to multimodal settings. Multimodal diffusion models~\citep{he2026vidlada,li2025sparse,tian2025mmada,chen2025unified,ji2025denoising,wen2025dvla,ye2025dreamvl,li2026omni,cheng2025sdar} are typically built upon pretrained dLLMs and incorporate visual representations to handle vision-language tasks. For example, methods such as LLaDA-V~\citep{you2025lladav}, LaViDa~\citep{li2025lavida}, and Dimple~\citep{yu2025dimple} align image features into the same embedding space as text tokens through visual instruction tuning, enabling joint modeling of visual and linguistic information. Another line of methods~\citep{yang2025mmada,swerdlow2025unified,shi2025muddit}, such as LLaDA-O~\citep{you2026llada}, Lumina-dimoo~\citep{xin2025lumina}, and LaViDa-O~\citep{li2025lavidao}, further combines fine-tuning and reinforcement learning to jointly model discretized visual and textual representations for unified multimodal generation and understanding. In addition, methods such as DiffusionVL~\citep{zeng2025diffusionvl} and Fudoki~\citep{wang2025fudoki} transform autoregressive models into the discrete diffusion paradigm through diffusion fine-tuning, providing another path for multimodal diffusion modeling.

\textbf{Decoding strategy in dLLM.} Despite the advantages of parallel generation and global modeling offered by discrete masked diffusion, the inference efficiency~\citep{oba2026stopping,nguyen2025attention,huang2025mask,gao2025self,wu2025free,wu2025fast,chen2025dparallel,ben2025accelerated} and generation quality of diffusion large language models largely depend on the decoding strategy. One line of work leverages cache reuse~\citep{li2025diffuspec,jiang2025d,wang2025diffusion,song2026sparse,ma2025dkv,liu2025dllm} and token pruning~\citep{long2026focus,xiao2026streaming,li2026dip,chen2025dpad,chang2026d3tom} to alleviate the computational bottlenecks introduced by global attention. Another line of research focuses on optimizing sampling~\citep{feng2026dvoting,xiang2026stop,chen2026dflash,li2026plan,dong2025saber,kong2025accelerating,wang2025creditdecoding,qihierarchy,wang2025time,kim2025klass}, early stopping~\citep{li2025diffusion,li2025beyond,chen2025beyond}, and step scheduling~\citep{luo2026dsb,zhang2026swordsman,fu2025bits,wei2025accelerating,fu2025bits,zhang2026swordsman,bao2025learning,luxembourg2025plan} to reduce redundant denoising iterations. Existing approaches for mitigating hallucinations in diffusion models include self-corrective decoding~\citep{hong2025wide,huang2025don,tian2025finish,chen2025optimizing,shu2026deferred,xin2025dmllm,zhong2026efficient} and prior constraints~\citep{zuo2026diffusion,cheng2026improving,wu2026time,seo2025fast,xiong2025unveiling,zhang2026learning,yang2026improving}, which improve model reliability by reducing error accumulation. Prior work replaces the binary mask-to-token transition with continuous token representations at each decoding step~\citep{ye2026rejection,zhu2025latent,zhong2026beyond,chen2026dmax,hersche2025soft,zheng2025continuously}. We examine hallucination in MDLMs from the perspective of step-wise uncertainty dynamics during denoising. Based on this observation, we propose StepRefine, a plug-and-play decoding method that mitigates hallucinations by detecting entropy spike steps and selectively applying latent refinement.

\section{Conclusion}

In this work, we extensively study hallucination in multimodal diffusion language models by looking at how predictive uncertainty evolves across denoising steps. We observe that during iterative denoising, MDLMs show sharp local increases in mean mask entropy at a small number of steps, which we call entropy spikes. These steps correlate with hallucinated tokens and mainly arise in the early-to-middle stages of denoising, where confidence-based decoding becomes less reliably grounded in the current context. Building on this observation, we introduce StepRefine, a training-free decoding strategy that detects entropy spikes via local trend deviation, switches to iterative latent refinement at those steps, and adaptively terminates once the predictive distribution stabilizes. Across four MDLM backbones and multiple benchmarks comprehensively covering general reasoning, hallucination, mathematical, and scientific reasoning, StepRefine obtains improved accuracy and can be combined with existing acceleration methods. In summary, our findings suggest that focusing refinement on a small number of high-uncertainty steps yields measurable improvements in MDLM decoding without additional training. 

\section*{Acknowledgments}
ZX was supported by the Australian Government Research Training Program (RTP) Scholarship.
This work was supported by the NVIDIA Singapore team through computational resources.

\clearpage
\bibliographystyle{plain}
\bibliography{neurips_2026}
\clearpage


\appendix

\section{Use of LLMs}
LLMs were used solely as writing and editing assistants in this study, with their role limited to improving grammar, spelling, clarity, and stylistic consistency. They did not contribute to the scientific content of the work, including research ideation, experimental design, data analysis, result interpretation, or any core intellectual contributions. All substantive content and scientific contributions are entirely attributable to the authors.
\section{Additional Experiments}
\subsection{Baseline}
\label{appendix:baseline}

All experiments are conducted on a single server with 8 NVIDIA H100 GPUs (80GB). To benchmark our method against recent training-free decoding strategies for diffusion-based models, we reproduce three representative baselines. All baselines are implemented under the same backbone and evaluation pipeline. We follow the original method designs and reported hyper-parameter settings whenever available to ensure a fair comparison.

\begin{itemize}
    \item \textbf{Wide-In, Narrow-Out (WINO)~\citep{hong2025wide}: }
WINO enables revokable decoding for DLLMs through a parallel draft-and-verify mechanism. At each decoding step, it first drafts multiple tokens using a lenient confidence threshold, then verifies previously decoded tokens with an auxiliary shadow block and re-masks low-confidence tokens for refinement. We follow the original semi-autoregressive decoding setting and use the reported drafting and verification thresholds.

    \item \textbf{Context Tokens are Anchors (CoTA)~\citep{zhao2026context}: }
CoTA mitigates cache-induced repetition in dMLLMs by preserving the information flow of context tokens. It enhances attention to nearby context tokens through a distance-aware intervention and adjusts decoding scores using deep-layer context-token entropy to suppress uncertain candidates. We integrate CoTA with the corresponding cache setting and follow the original hyper-parameter configuration.

    \item \textbf{Rejection Mixing (ReMix)~\citep{ye2026rejection}: }
ReMix accelerates DLLM decoding by introducing a continuous mixing state between masked and finalized token states. Undecoded positions are updated with probability-weighted embeddings to propagate semantic information, while unstable positions are rejected back to the masked state based on distributional divergence. We follow the original confidence threshold, mixing ratio, and rejection threshold settings.
\end{itemize}

We also examine the compatibility of our method with recent training-free inference acceleration methods for diffusion language models. These methods target different computational bottlenecks during diffusion decoding, such as feature reuse, KV-cache approximation, sparse attention, and delayed cache updates. Since our method only introduces an intra-step soft refinement before discrete token commitment, these acceleration methods can be applied during regular denoising steps and resumed after refinement.

\begin{itemize}
    \item \textbf{dLLM-Cache~\citep{liu2025dllm}: }
    dLLM-Cache accelerates diffusion language model inference by reusing intermediate features across adjacent denoising steps. It applies long-interval caching to static prompt tokens and adaptive short-interval updates to response tokens, where dynamically changing positions are selected according to value-vector similarity. When combined with our method, dLLM-Cache can be used during regular decoding steps, while soft-refinement steps are handled with an isolated temporary cache before refreshing the main cache after token commitment.

    \item \textbf{Fast-dLLM~\citep{wu2025fast}: }
    Fast-dLLM improves diffusion LLM inference through block-wise approximate KV caching and confidence-aware parallel decoding. Its cache mechanism reuses prefix, or prefix-and-suffix, KV states within a decoding block, while its parallel decoding strategy commits only high-confidence tokens to reduce dependency errors. In our setting, Fast-dLLM can operate during normal decoding, whereas soft refinement temporarily disables cache reuse and then returns the refined logits to the standard confidence-aware commitment rule.

    \item \textbf{SparseD~\citep{wang2025sparsed}: }
    SparseD reduces the quadratic attention cost in diffusion language models by using head-specific sparse attention patterns. It first applies full attention in early denoising steps, then pre-computes sparse query-key patterns and reuses them in later steps. Since SparseD modifies the attention computation rather than the discrete decoding trajectory, it is naturally compatible with our method; during soft refinement, one may either retain the sparse attention pattern or conservatively fall back to full attention.

    \item \textbf{dKV-Cache~\citep{ma2025dkv}: }
    dKV-Cache introduces a delayed KV-cache mechanism for diffusion language models by caching only decoded tokens whose representations have become relatively stable. To avoid instability immediately after token commitment, it delays cache insertion by one denoising step and periodically refreshes cached states. When integrated with our method, dKV-Cache remains active in regular denoising steps, while refinement uses a temporary soft-state cache; newly committed tokens after refinement follow the original one-step delayed caching rule.
\end{itemize}

\subsection{Benchmarks}

In this section, we briefly describe each benchmark used in our evaluation, grouped into four categories: general reasoning, hallucination, mathematical reasoning, and scientific reasoning.

\subsubsection{General Reasoning \& Understanding}
\begin{itemize}
    \item \textbf{V*Bench (vstar\_bench)~\citep{wu2024v}}:
    a multimodal benchmark built on diverse real-world images that probes visual understanding, spatial reasoning, and fine-grained attribute recognition in VLMs.

    \item \textbf{RealWorldQA~\citep{grok2beta2024}}:
    a collection of real-world photographs (many taken from vehicle viewpoints) paired with short-answer questions, targeting basic spatial understanding and everyday scene reasoning.

    \item \textbf{MMVP (MultiModal Visual Patterns)~\citep{tong2024eyes}}:
    300 carefully crafted questions over ``CLIP-blind'' image pairs, testing whether models can tell apart subtle visual-pattern differences that standard visual encoders typically miss.

    \item \textbf{MMEval-Pro~\citep{huang2025mmevalpro}}:
    a calibrated multimodal benchmark that augments existing multiple-choice datasets (e.g., MMMU, ScienceQA, MathVista) with perception and knowledge-anchor questions, forming question triplets that suppress text-only shortcuts.

    \item \textbf{VMCBench~\citep{zhang2025automated}}:
    a unified multiple-choice benchmark ($\sim$9k questions) that consolidates 20 VQA datasets via AutoConverter, enabling consistent and scalable evaluation of VLMs across diverse visual-linguistic scenarios.
\end{itemize}

\subsubsection{Hallucination Benchmark}
\begin{itemize}
    \item \textbf{MMHalu (MMHal-Bench)~\citep{sun2023aligning}}:
    a general-purpose hallucination benchmark built on diverse image-question pairs, measuring how often models generate visually inconsistent or fabricated content in realistic settings.

    \item \textbf{Bingo (Bias and Interference Challenges)~\citep{cui2023holistic}}:
    a diagnostic benchmark with image-text challenge sets that target bias-driven errors (region, OCR, factual) together with interference-induced hallucinations.

    \item \textbf{POPE (Polling-based Object Probing Evaluation)~\citep{li2023pope}}:
    an object-hallucination probe built from balanced yes/no questions about object existence, providing a stable, instruction-agnostic measure of image-faithful generation.
\end{itemize}

\subsubsection{Mathematical Reasoning}
\begin{itemize}
    \item \textbf{MathVision (MATH-Vision)~\citep{wang2024mathvision}}:
    3,040 visual math problems drawn from real competitions, spanning 16 disciplines and 5 difficulty levels, to assess multimodal mathematical reasoning in exam-style settings.

    \item \textbf{MathVista~\citep{lu2023mathvista}}:
    a unified benchmark of 6,141 examples from 28 existing datasets together with three new subsets (IQTest, FunctionQA, PaperQA), evaluating mathematical reasoning over diagrams, plots, and scientific figures.

    \item \textbf{MathVerse~\citep{zhang2024mathverse}}:
    2,612 diagram-based math problems expanded into $\sim$15k multimodal variants, enabling fine-grained evaluation of whether MLLMs actually reason over diagrams rather than rely on text alone.

    \item \textbf{VisuLogic~\citep{xu2025visulogic}}:
    1,000 single-choice visual reasoning problems across six categories (quantitative shifts, spatial relations, attribute comparisons, etc.), explicitly designed to rule out language-only shortcuts.

    \item \textbf{Geometry3K~\citep{lu2021inter}}:
    3,002 multiple-choice Euclidean geometry problems paired with diagrams and formalized structural annotations, covering high-school to competition-level difficulty.
\end{itemize}

\subsubsection{Scientific Reasoning}
\begin{itemize}
    \item \textbf{MMK12~\citep{meng2025mm}}:
    a K-12 multimodal reasoning dataset from the MM-Eureka framework; its evaluation split contains 500 multiple-choice questions per discipline (mathematics, physics, chemistry, biology), covering foundational scientific reasoning.
\end{itemize}

\subsection{Additional Benchmark Results}
\label{appendix:extra_results}

Table~\ref{tab:appendix_extra_results} reports StepRefine results on the remaining benchmarks that are not included in the main paper. The same trend holds: StepRefine consistently improves over the baseline across all backbones.

\begin{table}[h]
    \centering
    \small
    \setlength{\tabcolsep}{4pt}
    \renewcommand{\arraystretch}{1.4}
    \begin{tabular}{lc>{\columncolor{myhl}}cc>{\columncolor{myhl}}cc>{\columncolor{myhl}}cc>{\columncolor{myhl}}c}
        \toprule
        \multirow{2}{*}{\textbf{Benchmark}}
        & \multicolumn{2}{c}{\textbf{LLaDA-V-8B}}
        & \multicolumn{2}{c}{\textbf{LLaDA-O-8B}}
        & \multicolumn{2}{c}{\textbf{Dream-VL-7B}}
        & \multicolumn{2}{c}{\textbf{LaViDA-L-8B}} \\
        \cmidrule(lr){2-3} \cmidrule(lr){4-5} \cmidrule(lr){6-7} \cmidrule(lr){8-9}
        & Base & \cellcolor{myhl}\textbf{StepRefine}
        & Base & \cellcolor{myhl}\textbf{StepRefine}
        & Base & \cellcolor{myhl}\textbf{StepRefine}
        & Base & \cellcolor{myhl}\textbf{StepRefine} \\
        \midrule
        \multicolumn{9}{c}{\textit{\textbf{Hallucination Benchmarks}}} \\
        \midrule
        POPE-P        & 87.9 & \textbf{88.7}~\up{0.8} & 83.8 & \textbf{84.9}~\up{1.1} & 81.7 & \textbf{84.5}~\up{2.8} & 64.9 & \textbf{66.1}~\up{1.2} \\
        POPE-A        & 86.2 & \textbf{87.5}~\up{1.3} & 83.3 & \textbf{83.8}~\up{0.5} & 80.1 & \textbf{83.4}~\up{3.3} & 61.7 & \textbf{64.8}~\up{3.1} \\
        \midrule
        \multicolumn{9}{c}{\textit{\textbf{Mathematical Reasoning}}} \\
        \midrule
        MathVerse     & 34.1 & \textbf{35.3}~\up{1.2} & 38.6 & \textbf{42.6}~\up{4.0} & 31.4 & \textbf{34.9}~\up{3.5} & 32.1 & \textbf{33.9}~\up{1.8} \\
        Geometry3K    & 20.9 & \textbf{22.3}~\up{1.4} & 26.6 & \textbf{29.5}~\up{2.9} & 37.4 & \textbf{39.1}~\up{1.7} & 15.3 & \textbf{15.7}~\up{0.4} \\
        VisuLogic     & 24.5 & \textbf{27.7}~\up{3.2} & 35.2 & \textbf{36.3}~\up{1.1} & 24.8 & \textbf{27.1}~\up{2.3} & 23.1 & \textbf{25.9}~\up{2.8} \\
        MMK12-Math    & 38.8 & \textbf{42.0}~\up{3.2} & 26.8 & \textbf{28.0}~\up{1.2} & 42.8 & \textbf{45.6}~\up{2.8} & 36.2 & \textbf{39.0}~\up{2.8} \\
        \midrule
        \multicolumn{9}{c}{\textit{\textbf{Scientific Reasoning}}} \\
        \midrule
        MMK12-Bio     & 39.6 & \textbf{40.2}~\up{0.6} & 27.4 & \textbf{28.8}~\up{1.4} & 31.8 & \textbf{35.8}~\up{4.0} & 38.0 & \textbf{41.4}~\up{3.4} \\
        \bottomrule
    \end{tabular}
    \vspace{1em}
    \caption{Additional benchmark results of StepRefine across the four backbones. Accuracy is reported for all benchmarks.}
    \vspace{-2em}
    \label{tab:appendix_extra_results}
\end{table}

\subsection{Extensive ablations on model parameter setting}
\label{appendix:parameter_setting}

In this section, we provide additional ablations on decoding hyper-parameters that complement the LLaDA-V-8B results in Table~\ref{tab:ablation_steprefine_combined} of the main paper. We extend the remasking-strategy and step-budget studies to Dream-VL-7B (Table~\ref{tab:ablation_steprefine_combined_dream}), and further report a block-length study on LLaDA-V-8B and LLaDA-O-8B (Table~\ref{tab:ablation_blocklen}). All numbers are MMHalu scores in $[0, 6]$.

Table~\ref{tab:ablation_steprefine_combined_dream} mirrors the main-paper layout on Dream-VL-7B. (a) shows that StepRefine consistently improves the MMHalu score over the baseline under all three remasking strategies, with absolute gains ranging from $+2.3\%$ to $+3.7\%$. (b) varies the step budget $T_{\max}$ at two generation lengths $L\in\{128, 256\}$; StepRefine improves the score in every $(L, T_{\max})$ configuration tested. The trends agree with the LLaDA-V-8B results, indicating that the gains transfer across backbones.

\begin{table}[h]
    \centering
    \begin{minipage}[t]{0.36\textwidth}
        \centering
        \small
        \setlength{\tabcolsep}{3pt}
        \renewcommand{\arraystretch}{1.4}
        \begin{tabular}{lcc}
            \toprule
            \textbf{Strategy} & Base & \textbf{StepRefine} \\
            \midrule
            Random          & 2.73 & \textbf{2.95} \\
            Low Confidence  & 3.47 & \textbf{3.68} \\
            Top-$k$ Margin  & 3.25 & \textbf{3.39} \\
            \bottomrule
        \end{tabular}
        \\[3pt]
        {\footnotesize \textbf{(a) Remasking strategy}}
    \end{minipage}
    \hfill
    \begin{minipage}[t]{0.60\textwidth}
        \centering
        \small
        \setlength{\tabcolsep}{3pt}
        \renewcommand{\arraystretch}{1.26}
        \begin{tabular}{lcccccc}
            \toprule
            \multirow{2}{*}{\textbf{Method}}
            & \multicolumn{3}{c}{$L=256$} & \multicolumn{3}{c}{$L=128$} \\
            \cmidrule(lr){2-4} \cmidrule(lr){5-7}
            & $T{=}64$ & $T{=}128$ & $T{=}256$ & $T{=}32$ & $T{=}64$ & $T{=}128$ \\
            \midrule
            Base               & 3.37 & 3.54 & 3.47 & 3.18 & 3.69 & 3.58 \\
            \textbf{StepRefine} & \textbf{3.55} & \textbf{3.63} & \textbf{3.68} & \textbf{3.72} & \textbf{3.86} & \textbf{3.67} \\
            \bottomrule
        \end{tabular}
        \\[3pt]
        {\footnotesize \textbf{(b) Step budget at two generation lengths}}
    \end{minipage}
    \vspace{0.4cm}
    \caption{Ablations of decoding settings for Dream-VL-7B on MMHalu. (a) Baseline vs. StepRefine across remasking strategies. (b) Baseline vs. StepRefine across step budgets and generation lengths.}
    \label{tab:ablation_steprefine_combined_dream}
\end{table}

Table~\ref{tab:ablation_blocklen} reports MMHalu scores under four block-length settings on LLaDA-V-8B and LLaDA-O-8B. StepRefine consistently improves over the baseline at every block length on both backbones, with the largest absolute gain on LLaDA-O-8B at block length $32$ ($+4.7\%$).

\begin{table}[h]
    \centering
    \small
    \setlength{\tabcolsep}{6pt}
    \renewcommand{\arraystretch}{1.4}
    \begin{tabular}{ccccc}
        \toprule
        \multirow{2}{*}{\textbf{Block Length}}
        & \multicolumn{2}{c}{\textbf{LLaDA-V-8B}}
        & \multicolumn{2}{c}{\textbf{LLaDA-O-8B}} \\
        \cmidrule(lr){2-3} \cmidrule(lr){4-5}
        & Base & \textbf{StepRefine}
        & Base & \textbf{StepRefine} \\
        \midrule
        32   & 3.22 & \textbf{3.29} & 3.63 & \textbf{3.91} \\
        64   & 3.29 & \textbf{3.40} & 3.61 & \textbf{3.67} \\
        128  & 3.37 & \textbf{3.46} & 3.73 & \textbf{3.81} \\
        256  & 3.35 & \textbf{3.50} & 3.76 & \textbf{4.01} \\
        \bottomrule
    \end{tabular}
    \vspace{0.4cm}
    \caption{Ablation of block length on MMHalu across four block-length settings on the LLaDA-V-8B and LLaDA-O-8B backbones. }
    \label{tab:ablation_blocklen}
\end{table}

\subsection{Compatibility with other acceleration methods}
\label{appendix:sparsed}

We additionally report the compatibility of StepRefine with SparseD~\citep{wang2025sparsed}, which differs from the cache-based accelerators in the main paper by reducing attention cost through head-specific sparse patterns. As shown in Table~\ref{tab:accel_compat_sparsed}, SparseD + StepRefine preserves SparseD's speedup and improves accuracy on every benchmark, indicating that the orthogonality of StepRefine extends beyond cache-based methods to attention-pattern-based ones.

\begin{table}[h]
    \centering
    \small
    \setlength{\tabcolsep}{4pt}
    \renewcommand{\arraystretch}{1.4}
    \begin{tabular}{lcccccc}
        \toprule
        \multirow{2}{*}{\textbf{Benchmark}}
        & \multicolumn{3}{c}{\textbf{SparseD}}
        & \multicolumn{3}{c}{\textbf{SparseD + StepRefine}} \\
        \cmidrule(lr){2-4} \cmidrule(lr){5-7}
        & Acc & Speedup & Tput
        & Acc & Speedup & Tput \\
        \midrule
        VStar  & 56.0 & \spd{1.15} & 3.17 & \textbf{58.6} & \spd{1.18} & 3.25 \\
        MMVP   & 39.3 & \spd{1.39} & 9.08 & \textbf{40.7} & \spd{1.42} & 9.29 \\
        MMHalu & 3.37 & \spd{1.38} & 3.76 & \textbf{3.46} & \spd{1.42} & 3.88 \\
        Bingo  & 2.41 & \spd{1.28} & 4.28 & \textbf{2.42} & \spd{1.30} & 4.33 \\
        \bottomrule
    \end{tabular}
    \vspace{0.4cm}
    \caption{Compatibility of StepRefine with SparseD across four benchmarks. Speedups are measured relative to the unaccelerated LLaDA-V baseline.}
    \label{tab:accel_compat_sparsed}
\end{table}

\vspace{-2em}
\section{Visualizations}
\label{app:visualization}

\subsection{Additional Examples of Entropy Spike Detection}

\begin{figure}[t]
  \centering
  \vspace{1em}
  \includegraphics[width=1\linewidth]{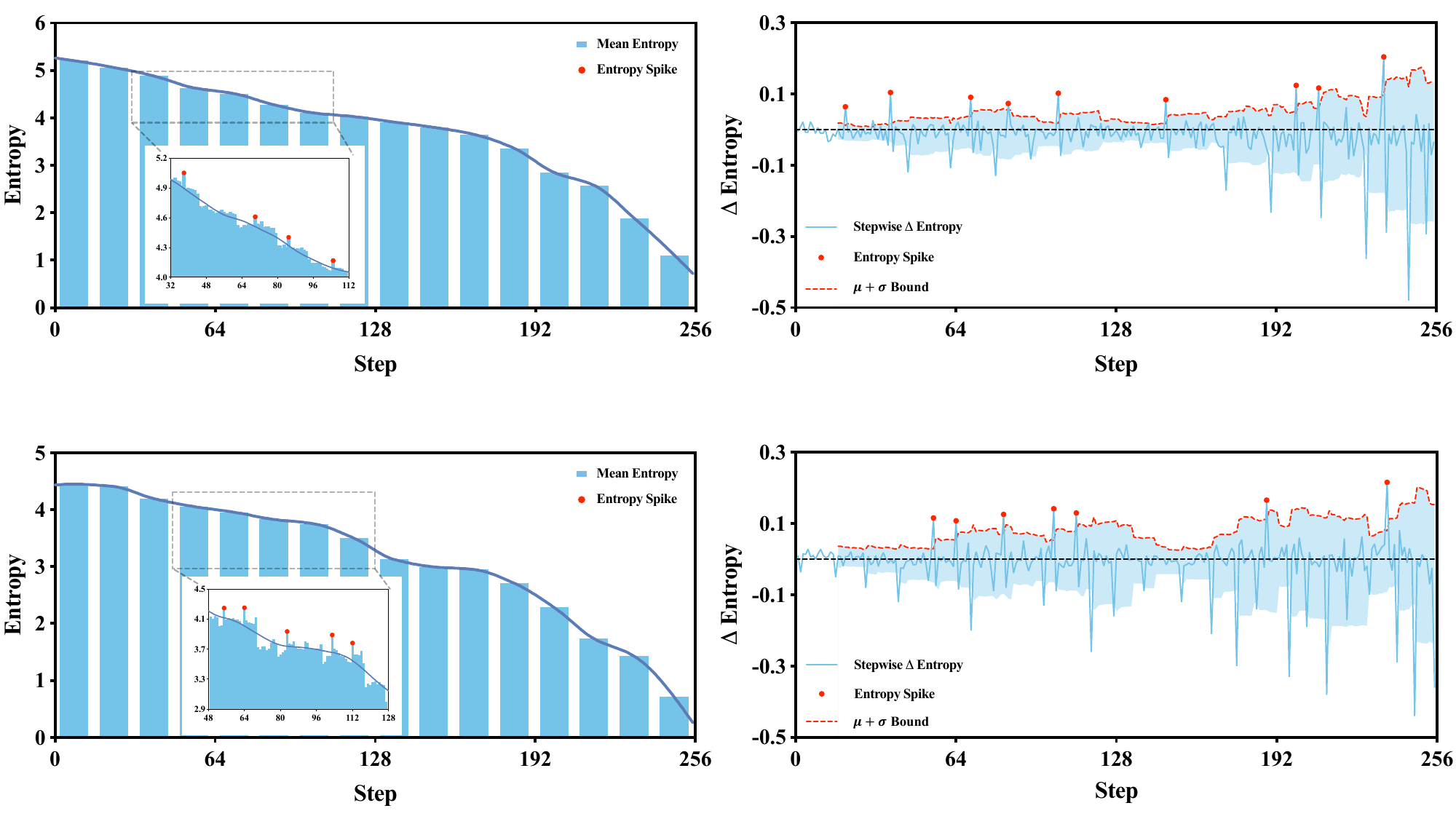}
  \caption{Additional examples of entropy spike detection on two samples. We plot the mean mask entropy across denoising steps and mark the steps flagged as entropy spikes by the $\mu+\sigma$ threshold over the local calibration window.}
  \label{fig:appendix_spike_detection}
\end{figure}

To complement Figure~\ref{fig:main1}, we provide two additional samples in Figure~\ref{fig:appendix_spike_detection}. Each sample shows the mean mask entropy across denoising steps, with the detected entropy spikes marked along the trajectory.

\subsection{General and Domain-specific Multimodal Reasoning Visualization}

\begin{figure}[t]
  \centering
  \vspace{1em}
  \includegraphics[width=1\linewidth]{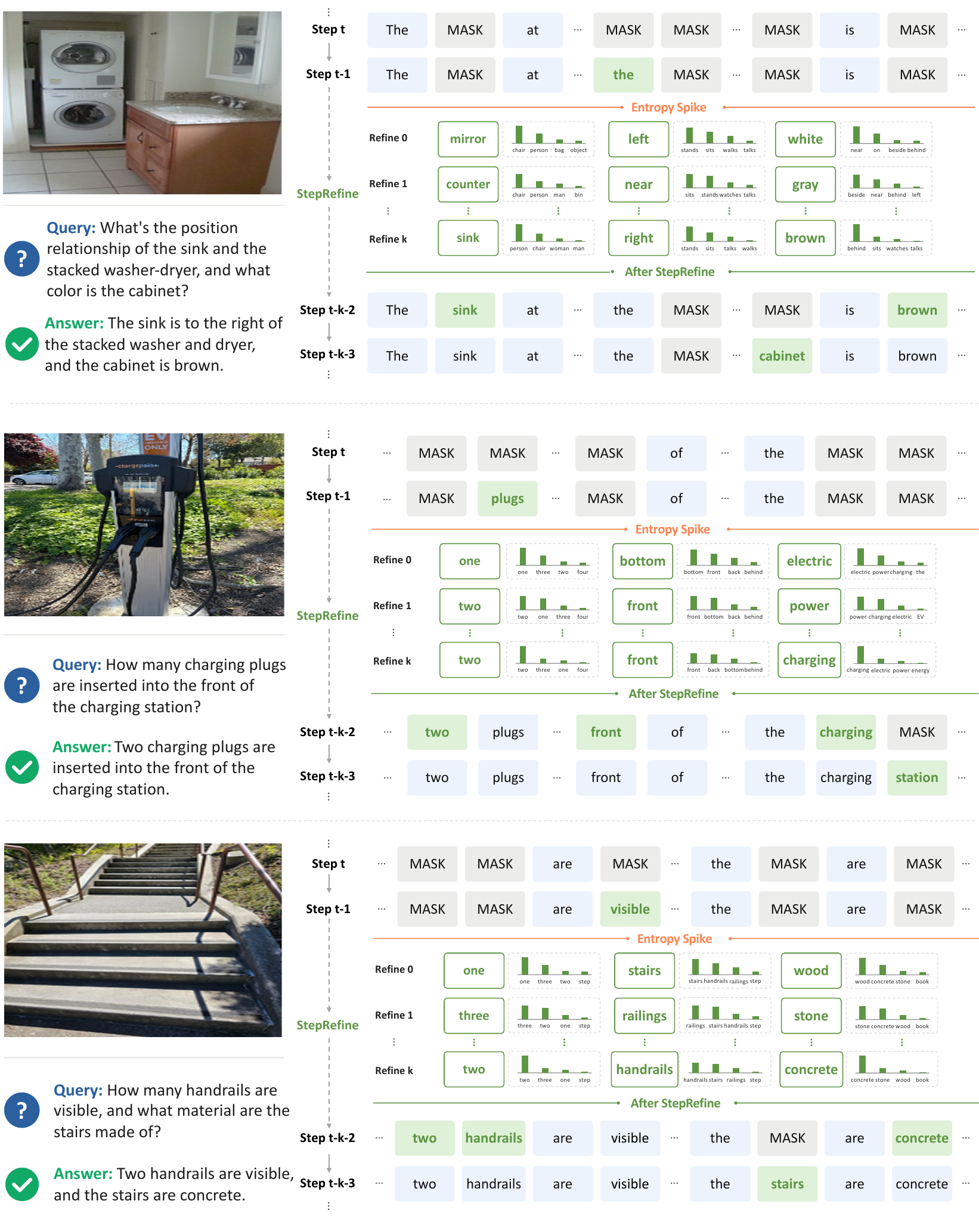}
\caption{Example visualization of StepRefine on general reasoning \& hallucination benchmarks. }
  \label{fig:ablation_general_visualization}
\end{figure}

\begin{figure}[t]
  \centering
  \vspace{1em}
  \includegraphics[width=1\linewidth]{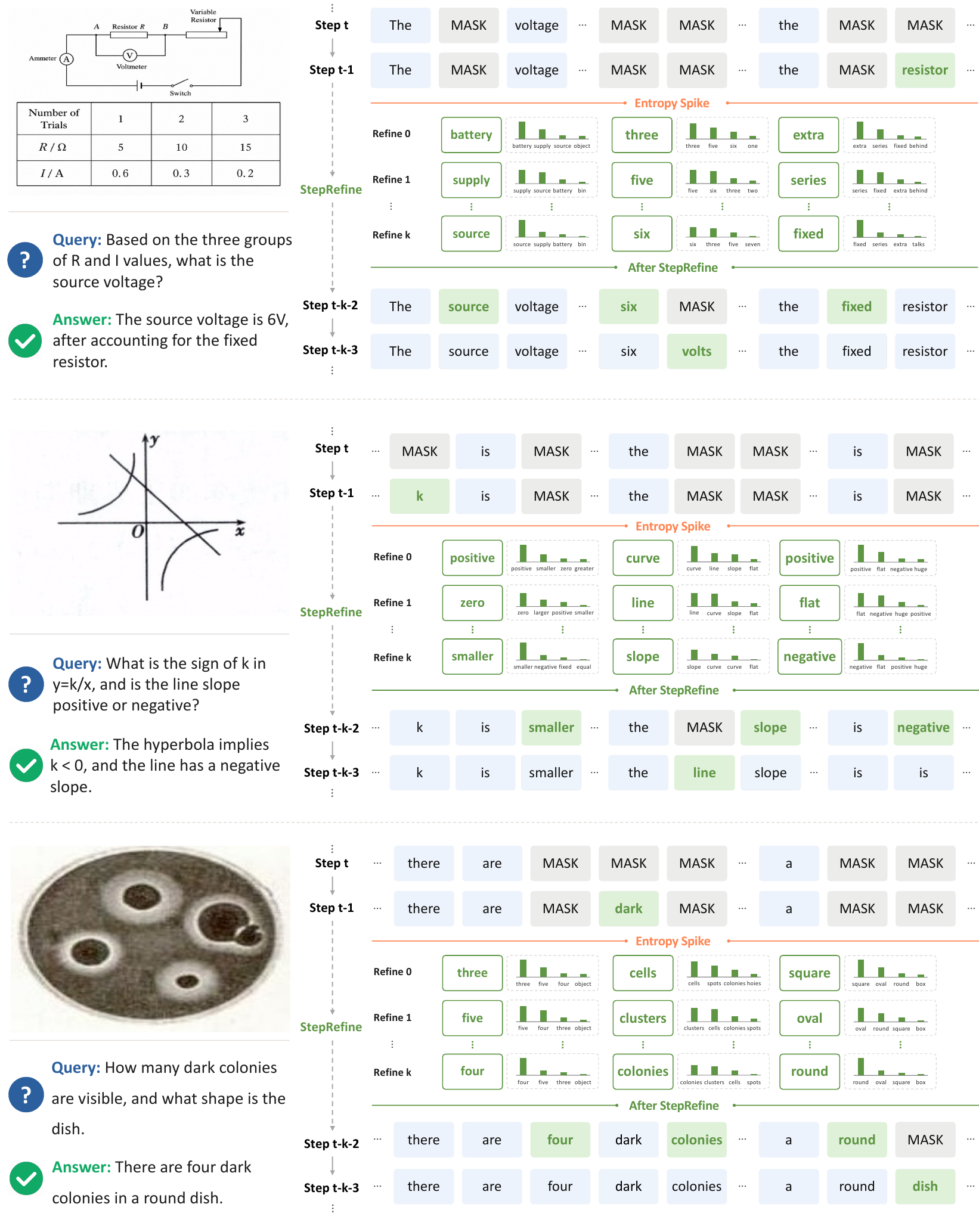}
\caption{Example visualization of StepRefine on mathematical \& scientific benchmarks. }
  \label{fig:ablation_specific_visualization}
\end{figure}

Figure~\ref{fig:ablation_general_visualization} presents more qualitative examples of StepRefine's decoding trajectories on general reasoning and hallucination benchmarks, covering a broader range of question types and visual scenes. For domain-specific settings, additional decoding visualizations on mathematical and scientific reasoning benchmarks are shown in Figure~\ref{fig:ablation_specific_visualization}.

\subsection{Challenge in Specific Scenarios}

\begin{figure}[t]
  \centering
  \vspace{1em}
  \includegraphics[width=1\linewidth]{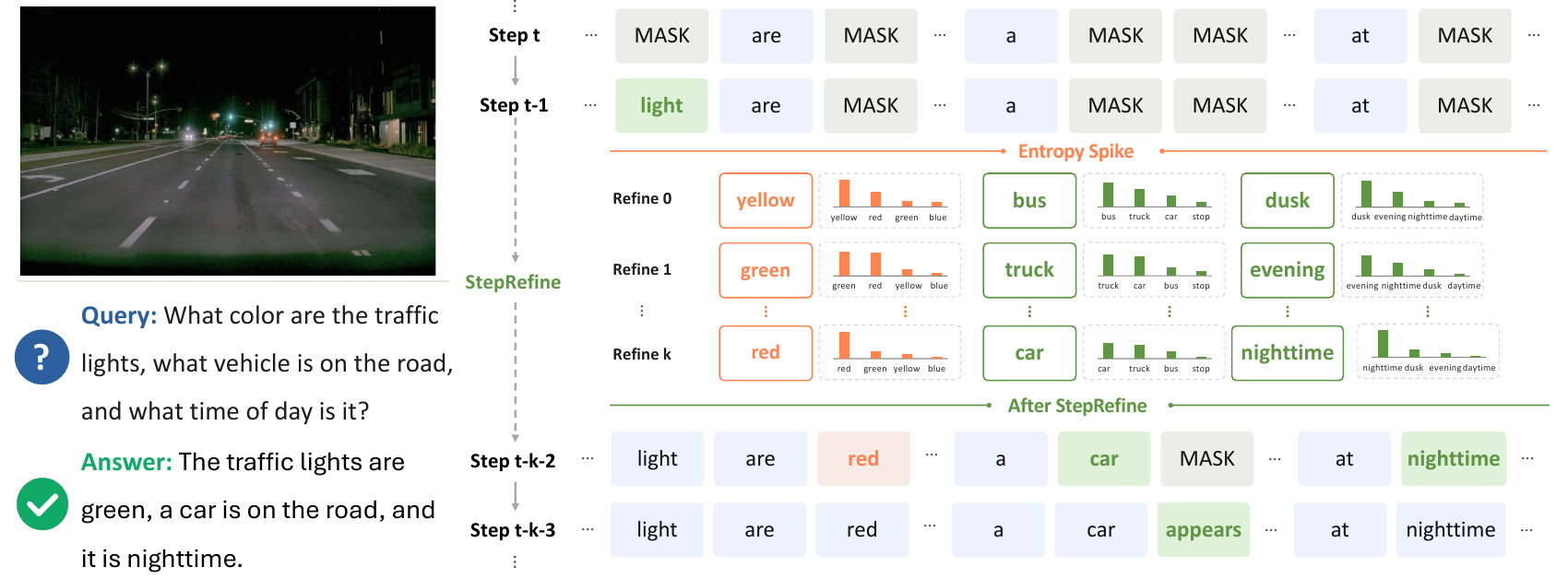}
\caption{Example visualization of StepRefine on a challenging case where intra-step refinement fails to recover the correct answer. }
  \label{fig:ablation_error_visualization}
\end{figure}

We further include a case in Figure~\ref{fig:ablation_error_visualization}, where intra-step refinement is unable to recover the correct answer in particularly challenging scenarios. 

\section{Broader Impacts}

All experiments are conducted on publicly released models and benchmarks; no personal or sensitive data are collected or inferred, and all qualitative examples are drawn from publicly available images. Our approach does not involve human-subject interaction and does not raise specific ethical concerns beyond those typically associated with AI model development and evaluation.


\clearpage
\newpage
\section*{NeurIPS Paper Checklist}

\begin{enumerate}

\item {\bf Claims}
    \item[] Question: Do the main claims made in the abstract and introduction accurately reflect the paper's contributions and scope?
    \item[] Answer: \answerYes{} 
    \item[] Justification: Our contributions are summarized in the Abstract and Introduction sections.

\item {\bf Limitations}
    \item[] Question: Does the paper discuss the limitations of the work performed by the authors?
    \item[] Answer: \answerYes{} 
    \item[] Justification: We discuss failure cases in Appendix. 

\item {\bf Theory assumptions and proofs}
    \item[] Question: For each theoretical result, does the paper provide the full set of assumptions and a complete (and correct) proof?
    \item[] Answer: \answerNA{} 
    \item[] Justification: This paper does not include theoretical results. 

    \item {\bf Experimental result reproducibility}
    \item[] Question: Does the paper fully disclose all the information needed to reproduce the main experimental results of the paper to the extent that it affects the main claims and/or conclusions of the paper (regardless of whether the code and data are provided or not)?
    \item[] Answer: \answerYes{} 
    \item[] Justification: Please see the setup in Sec.4.1 and detailed experimental settings in appendix B. 

\item {\bf Open access to data and code}
    \item[] Question: Does the paper provide open access to the data and code, with sufficient instructions to faithfully reproduce the main experimental results, as described in supplemental material?
    \item[] Answer: \answerYes{} 
    \item[] Justification: The code will be released upon acceptance.

\item {\bf Experimental setting/details}
    \item[] Question: Does the paper specify all the training and test details (e.g., data splits, hyperparameters, how they were chosen, type of optimizer) necessary to understand the results?
    \item[] Answer: \answerYes{} 
    \item[] Justification: The detailed results can be found in Appendix. 

\item {\bf Experiment statistical significance}
    \item[] Question: Does the paper report error bars suitably and correctly defined or other appropriate information about the statistical significance of the experiments?
    \item[] Answer: \answerYes{} 
    \item[] Justification: The detailed results can be found in Appendix.

\item {\bf Experiments compute resources}
    \item[] Question: For each experiment, does the paper provide sufficient information on the computer resources (type of compute workers, memory, time of execution) needed to reproduce the experiments?
    \item[] Answer: \answerYes{} 
    \item[] Justification: The detailed results can be found in Appendix.
    
\item {\bf Code of ethics}
    \item[] Question: Does the research conducted in the paper conform, in every respect, with the NeurIPS Code of Ethics \url{https://neurips.cc/public/EthicsGuidelines}?
    \item[] Answer: \answerYes{} 
    \item[] Justification: We have read the NeurIPS Code of Ethics and confirm that our research complies with these ethical standards. 

\item {\bf Broader impacts}
    \item[] Question: Does the paper discuss both potential positive societal impacts and negative societal impacts of the work performed?
    \item[] Answer: \answerYes{} 
    \item[] Justification: Please see the Appendix.
    
\item {\bf Safeguards}
    \item[] Question: Does the paper describe safeguards that have been put in place for responsible release of data or models that have a high risk for misuse (e.g., pre-trained language models, image generators, or scraped datasets)?
    \item[] Answer: \answerNA{} 
    \item[] Justification: This paper poses no such risks.

\item {\bf Licenses for existing assets}
    \item[] Question: Are the creators or original owners of assets (e.g., code, data, models), used in the paper, properly credited and are the license and terms of use explicitly mentioned and properly respected?
    \item[] Answer: \answerYes{} 
    \item[] Justification: All the thing in this paper credited and are the license and terms of use explicitly mentioned and properly respected. 

\item {\bf New assets}
    \item[] Question: Are new assets introduced in the paper well documented and is the documentation provided alongside the assets?
    \item[] Answer: \answerNA{} 
    \item[] Justification: This paper does not release new assets.

\item {\bf Crowdsourcing and research with human subjects}
    \item[] Question: For crowdsourcing experiments and research with human subjects, does the paper include the full text of instructions given to participants and screenshots, if applicable, as well as details about compensation (if any)? 
    \item[] Answer: \answerNA{} 
    \item[] Justification: This paper does not involve crowdsourcing nor research with human subjects.

\item {\bf Institutional review board (IRB) approvals or equivalent for research with human subjects}
    \item[] Question: Does the paper describe potential risks incurred by study participants, whether such risks were disclosed to the subjects, and whether Institutional Review Board (IRB) approvals (or an equivalent approval/review based on the requirements of your country or institution) were obtained?
    \item[] Answer: \answerNA{} 
    \item[] Justification: This paper does not involve crowdsourcing nor research with human subjects.

\item {\bf Declaration of LLM usage}
    \item[] Question: Does the paper describe the usage of LLMs if it is an important, original, or non-standard component of the core methods in this research? Note that if the LLM is used only for writing, editing, or formatting purposes and does \emph{not} impact the core methodology, scientific rigor, or originality of the research, declaration is not required.
    \item[] Answer: \answerNA{} 
    \item[] Justification: The method development in this paper does not involve LLMs as any components.

\end{enumerate}

\end{document}